\documentclass[11pt]{article}

\usepackage[final]{acl}

\usepackage{colortbl}
\usepackage{xcolor}

\newcommand{\degrade}[1]{\cellcolor{red!#1}}
\usepackage{times}
\usepackage{latexsym}
\usepackage[ruled,vlined,linesnumbered]{algorithm2e}
\usepackage{amsfonts}
\usepackage[T1]{fontenc}

\usepackage[utf8]{inputenc}

\usepackage{microtype}

\usepackage{inconsolata}

\usepackage{graphicx}
\usepackage{amsmath}

\usepackage{booktabs, multirow}
\usepackage{subcaption} 
\usepackage{tcolorbox}
\tcbuselibrary{breakable, skins}
\usepackage{enumitem}

\title{Live-Evo: Online Evolution of Agentic Memory from Continuous Feedback}

\author{
\textbf{Yaolun Zhang\textsuperscript{1,2,*}} \quad
\textbf{Yiran Wu\textsuperscript{2,3,*}} \quad
Yijiong Yu\textsuperscript{1} \quad
Qingyun Wu\textsuperscript{2,3} \quad
Huazheng Wang\textsuperscript{1,2}
\\[0.5em]
\textsuperscript{1}Oregon State University \quad
\textsuperscript{2}AG2 AI \quad
\textsuperscript{3}Penn State University
\\[0.5em]
\texttt{\{zhanyaol,yuyiji,huazheng.wang\}@oregonstate.edu, ykw5399@psu.edu, qingyun@ag2.ai} 
\\[0.5em]
\textsuperscript{*}Equal contribution. \quad
}

\begin{document}
\maketitle
\begin{abstract}
Large language model (LLM) agents are increasingly equipped with memory, which are stored experience and reusable guidance that can improve task-solving performance. Recent \emph{self-evolving} systems update memory based on interaction outcomes, but most existing evolution pipelines are developed for static train/test splits and only approximate online learning by folding static benchmarks, making them brittle under true distribution shift and continuous feedback. We introduce \textsc{Live-Evo}, an online self-evolving memory system that learns from a stream of incoming data over time. \textsc{Live-Evo} decouples \emph{what happened} from \emph{how to use it} via an Experience Bank and a Meta-Guideline Bank, compiling task-adaptive guidelines from retrieved experiences for each task. To manage memory online, \textsc{Live-Evo} maintains experience weights and updates them from feedback: experiences that consistently help are reinforced and retrieved more often, while misleading or stale experiences are down-weighted and gradually forgotten, analogous to reinforcement and decay in human memory. On the live \textit{Prophet Arena} benchmark over a 10-week horizon, \textsc{Live-Evo} improves Brier score by 20.8\% and increases market returns by 12.9\%, while also transferring to deep-research benchmarks with consistent gains over strong baselines. Visit our website for more details: \url{https://ag2ai.github.io/live-evo-page/}.

\end{abstract}

\section{Introduction}

Large Language Models (LLMs) have increasingly been adopted as the backbone of agent systems, enabling agents to interact with external environments through tool usage and to solve complex, multi-step tasks \cite{wu2023autogenenablingnextgenllm,zhang2024pybenchevaluatingllmagent, wu2025excytin, wu2024stateflow}. 
Recent work has proposed self-evolving agents 
\cite{gao2025surveyselfevolvingagentspath,qiu2025alitageneralistagentenabling,long2025seeinglisteningrememberingreasoning}, which allow agents to learn from a training set by constructing tools, knowledge, and task-solving strategies to better accomplish the tasks. Specifically, the knowledge and strategies learnt from past experiences are being recognized as \texttt{memory} of agents~\cite{jiang2024long, zhang2025survey}. These memory systems are typically organized into multiple levels, ranging from raw observations and interaction logs~\cite{park2023generativeagentsinteractivesimulacra,zhang2025survey,zhong2023memorybankenhancinglargelanguage} to higher-level summarized experiences and abstract guidelines~\cite{zhang2025memevolvemetaevolutionagentmemory,xu2025amemagenticmemoryllm}. 
During training, the agent can dynamically add, update, or remove memory entries based on its interaction outcomes.\cite{chhikara2025mem0buildingproductionreadyai} At test time, the agent leverages the evolved memory to guide decision-making on unseen tasks. These agents equipped with memory learned consistently outperform agents without memory evolution.

At the same time, memory evolution is inherently an online problem. In realistic deployments, an agent’s experience accrues sequentially, and its memory must be updated continually by adding new evidence, revising outdated entries, and consolidating recurring patterns, rather than being rebuilt from a static corpus. This perspective is closely related to classic online and continual learning in traditional machine learning~\cite{hoi2021online}, though the mechanisms and objectives for agent memory can differ. This shift raises a fundamental question: \textbf{how can LLM agents evolve continuously as new data arrives?}

\begin{figure}
    \centering
    \includegraphics[width=1\linewidth]{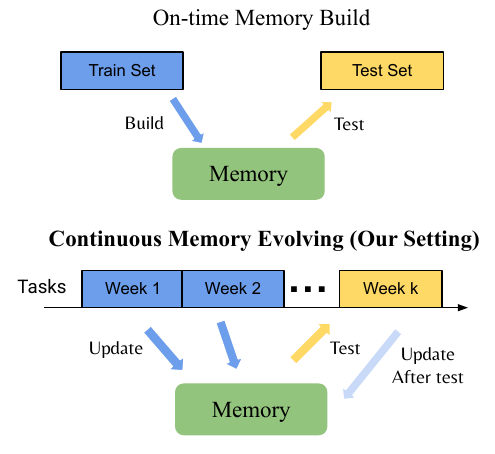}
    \caption{Traditional Self-Evolving Memory System build memory on training dataset and test with the evolved memory. While Live Self-Evolving Memory System build and learn to utilize Memory to tackle continuously new data. }
    \label{fig:compare_method}
\end{figure}

Live benchmarks like \textit{Prophet Arena}~\cite{yang2025llmasaprophetunderstandingpredictiveintelligence} and \textit{FutureX}~\cite{zeng2025futurexadvancedlivebenchmark} exemplify this paradigm by reframing agent evaluation as a longitudinal future prediction problem. In these benchmarks, agents are required to forecast probabilities of upcoming events, and are evaluated using both calibration-based metrics (e.g., Brier scores) and decision-oriented outcomes such as real market returns. In contrast to static retrieval or reasoning tasks, future prediction needs the agent to evolving during test time and continue adapt the memory to totally new tasks.
Figure \ref{fig:compare_method} shows the difference between traditional self-evolving memory and live self-evolving memory.

Only a few existing methods study memory for online task streams~\cite{wei2025evomemorybenchmarkingllmagent, wang2024agent}. However, they approximate “online” learning by splitting static benchmarks into folds, and therefore largely ignore distribution shift in truly streaming tasks. In contrast, live prediction benchmarks sample tasks from the real world, where environments and markets continually change over time. In this setting, success depends less on retrieving more information and more on \textit{judiciously leveraging past experience over time}. Past experience can provide useful inductive bias, but it can also become stale or misleading as patterns drift or break. Therefore, a self-evolving memory system must go beyond storage: it should actively curate experiences and learn when and how to use them.

We introduce \textsc{Live-Evo}, a self-evolving agentic memory system designed for continuous task streams. \textsc{Live-Evo} learns not only \textit{what happened before}, but also \textit{how to use experience} by maintaining an \textit{Experience Bank} and a \textit{Meta-Guideline Bank}. For each incoming task, the agent executes a four-stage loop. \textbf{Retrieve}: it generates search queries to retrieve relevant question--experience pairs. \textbf{Compile}: it compiles retrieved experiences into a task-specific guideline, instructed by \textit{Meta-Guidelines} that encode meta-heuristics for combining historical insights with the current task. \textbf{Act}: it performs \textit{ContrastiveEval} by producing and comparing two independent predictions, one guided by the compiled guideline and one as a memory-free baseline, to quantify the contribution of the guideline. \textbf{Update}: it updates experience weights based on the observed performance gap; if the guideline underperforms, the agent generates a new entry for the Meta-Guideline Bank. Finally, to control memory growth, \textsc{Live-Evo} summarizes trajectories from poorly solved cases into candidate experiences and commits them to the Experience Bank only after re-evaluation confirms an improvement.


We evaluate \textsc{Live-Evo} on the \textit{Prophet Arena}~\cite{yang2025llmasaprophetunderstandingpredictiveintelligence} benchmark over a 10-week horizon. Our results demonstrate that \textsc{Live-Evo} significantly outperforms static baselines, achieving a \textbf{20.8\%} improvement in Brier Score and a \textbf{12.9\%} increase in market returns. Furthermore, \textsc{Live-Evo} exhibits strong generalization on traditional deep-research benchmarks~\cite{chen2025xbenchtrackingagentsproductivity}, outperforming specialized state-of-the-art methods. Our ablation studies further confirm that the components are essential for maintaining performance in online benchmarks.

\begin{figure*}[t!]
    \centering
    \includegraphics[trim=3cm 2cm 5cm 2cm, clip,width=1\linewidth]{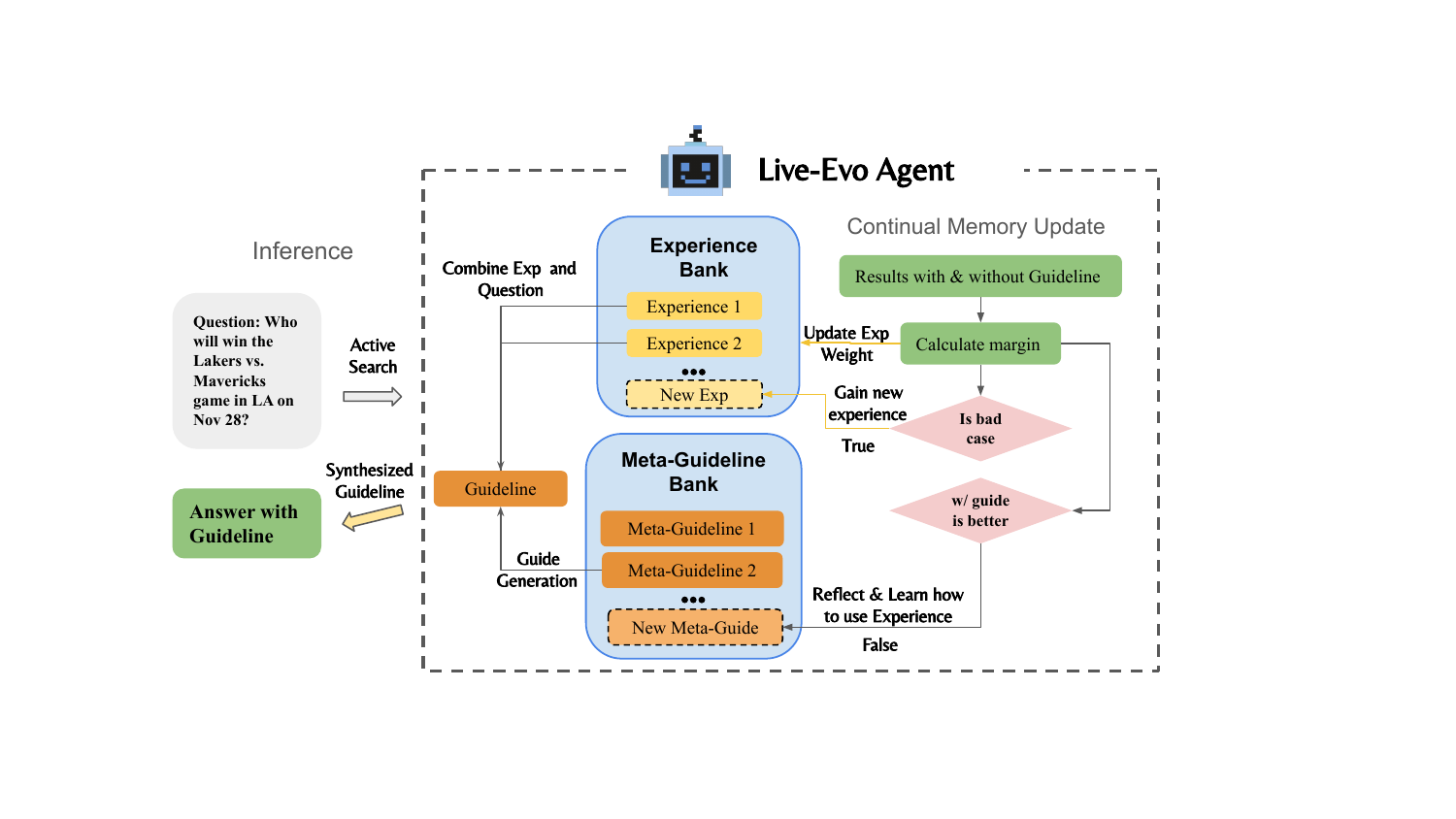}
    \caption{Structure of Live-Evo Agent. Given a question, the Live-Evo Agent will first search relevant experiences and generate a guideline based on the experiences, current task. Also, the generation will augmented by the meta-guideline bank, which teaches the agent how to combine experiences with current task. Inside the agent, the memory update mechanism continually updating experiences' weights and verifying new experiences and meta-guidelines. }
    \label{fig:method}
\end{figure*}

\section{Related Work}

\subsection{Self-Evolving Agentic Memory Systems}
Memory transforms LLM agents into persistent, adaptive systems capable of long-horizon task-solving~\cite{shan2025cognitive,qian2024experiential,yan2024dependingshouldmentoringllm} and life-long learning~\cite{zheng2025lifelonglearninglargelanguage,wang2024voyager}. Existing solutions follow a hierarchical evolution. Early methods~\cite{zheng2024synapse}, rely on static retrieval. While effective for repetitive tasks, they suffer from experience drift in changing environments~\cite{hu2025memory}.
To address this, recent memory systems support high-level memory operations including forgetting~\cite{liang2025selfevolvingagentsreflectivememoryaugmented,zhong2024memorybank}, building knowledge networks\cite{xu2025amemagenticmemoryllm}, and introducing heterogeneous memory structures~\cite{chhikara2025mem0}. Furthermore, some tasks focus on learn experience for task solving, \textsc{BoT}~\cite{yang2024buffer} synthesize high-level heuristics from past trajectories, \textsc{ExpeL}~\cite{zhao2024expelllmagentsexperiential} turns past trajectories into reusable experiences, and Agent Workflow Memory~\cite{wang2024agentworkflowmemory} record the action sequence of successful task. However, none of them focus on evolving on live benchmarks.


\subsection{Live Benchmarks}
Traditional static benchmarks are released at a fixed time and evaluate models on a closed dataset~\cite{wei2025browsecompsimplechallengingbenchmark,zhang2024pybenchevaluatingllmagent}. However, such benchmarks inevitably suffer from data leakage over time. To address this issue, several evaluation frameworks adopt a live setting by continuously introducing new tasks to assess models’ general capabilities~\cite{2023opencompass,xu2023supercluecomprehensivechineselarge}. Other approaches rely on live human feedback for evaluation~\cite{chiang2024chatbotarenaopenplatform}. More recent live benchmarks release new tasks over time, providing a continuous evaluation stream for specific tasks. For example, LiveCodeBench~\cite{jain2024livecodebenchholisticcontaminationfree} regularly releases new coding problems each quarter. Emerging \textit{future prediction benchmarks}, such as \textit{Prophet Arena}~\cite{yang2025llmasaprophetunderstandingpredictiveintelligence} and \textit{FutureX}~\cite{zeng2025futurexadvancedlivebenchmark}, introduce real-world tasks on a weekly basis, offering an ideal testbed for self-evolving agents.



\section{Method}




We introduce \textsc{Live-Evo}, an agentic memory system explicitly designed for true live benchmarks, where tasks arrive sequentially and feedback is revealed over time (See Figure~\ref{fig:method}). In contrast to prior memory systems that primarily store experiences and abstract them into static summaries, \textsc{Live-Evo} continuously optimizes how past experience is used over time, updating its memory policies as new data arrives and grounding each update in ongoing environmental feedback.

\textsc{Live-Evo} composes of two memory banks: an \emph{Experience Bank}  $\mathcal{E}$ and a \emph{Meta-Guideline Bank}  $\mathcal{M}$. The Experience Bank stores past task interactions in a structured, reusable form. When queried, the agent does not simply append retrieved trajectories into the prompt; instead, it applies a learned \emph{procedure} that distills retrieved experiences into task-relevant signals and actionable guidance, making the memory system's inductive bias explicit. Complementarily, the Meta-Guideline Bank stores higher-level \emph{composition instructions}, which meta-guidelines that specify how to transform retrieved experiences into a task-adaptive guideline under different conditions. Together, these two banks separate \emph{what happened before} (experience) from \emph{how to use it} (guideline), enabling memory usage to improve over time as new tasks arrive.

\begin{algorithm*}[t]
\small
\caption{\textsc{Live-Evo}}
\label{alg:live_evo}
\SetAlgoLined
\DontPrintSemicolon

\KwIn{Task stream batch $\mathcal{Q}$; Experience bank $\mathcal{E}$ with weights $\{w_e\}$; Meta-guideline bank $\mathcal{M}$; Bad-case fraction $\rho$}
\KwOut{Updated $\mathcal{E},\mathcal{M}$}

\BlankLine
\ForEach{$q \in \mathcal{Q}$}{
    $E_q,\ \hat m \leftarrow \textsc{Retrieve}(q,\mathcal{E},\mathcal{M})$ \tcp*[r]{\textbf{Retrieve:} top-$k$ experiences + selected meta-guideline}
    
    $g_q \leftarrow \textsc{CompileGuideline}(q,E_q,\hat m)$ \tcp*[r]{\textbf{Compile:} LLM produces task-specific memory guideline}
    
    $r^{\text{on}}_q,\ r^{\text{off}}_q,\ \tau_q \leftarrow \textsc{ContrastiveEval}(q,g_q)$ \tcp*[r]{\textbf{Act:} scores w/ and w/o memory; keep memory-on trajectory $\tau_q$}

    \tcp{\textbf{Update}}
    $W_{E_q} \leftarrow \textsc{UpdateWeights}(W_{E_q}, r^{\text{on}}_q - r^{\text{off}}_q)$ \tcp*[r]{ update weights of selected experiences} 
    
    \If{$r^{\text{on}}_q - r^{\text{off}}_q \le 0$}{$\mathcal{M} \leftarrow \mathcal{M}\ \cup\ \{\textsc{Reflect}(q,g_q, E_q)\}$\tcp*[r]{add new meta-guideline on failure}}

}

\tcp{\textbf{Update}}
$\mathcal{Q}_{\text{bad}} \leftarrow \textsc{SelectWorst}(\mathcal{Q},\{r^{\text{on}}_q\},\rho)$ \tcp*[r]{worst $\rho$ fraction of tasks solved with memory}
\ForEach{$q \in \mathcal{Q}_{\text{bad}}$}{
    $e^{\text{new}}_q \leftarrow \textsc{Summarize}(q,\tau_q)$ \tcp*[r]{summarize new experience from stored memory-on trajectory}
    \If{$\textsc{Eval}(q,e^{\text{new}}_q) > r^{\text{on}}_q$}{$\mathcal{E} \leftarrow \mathcal{E}\ \cup\ \{e^{\text{new}}_q\}$\tcp*[r]{re-evaluate with new experience and commit if it improves}}
}

\BlankLine
\Return{$\mathcal{E},\mathcal{M}$}\;
\end{algorithm*}

We formalize the self-evolving agentic memory system as a closed-loop decision process over the memory banks. For each new task batch, the agent operates through four stages:
\[
 \{\textsc{Retrieve}, \textsc{Compile}, \textsc{Act}, \textsc{Update}\}.
\]

Given a task, the agent first actively search its own memory to retrieve relevant experiences and the meta guideline. Then the agent compiles a guideline based on the meta instruction, the retrieved experiences and the task. The agent then solves the task with the compiled guideline. Finally, the trajectory and result of solving this task will be used to update the memory, including the experience bank and the meta guideline bank. Next we explain each stage of \textsc{Live-Evo} in detail (also see Algorithm~\ref{alg:live_evo}).

\subsection{Retrieve}
Given a task $q$, the agent $A$ first retrieves potentially relevant experiences and also a meta guidelines:
\[
E_q,\ \hat m \leftarrow \textsc{Retrieve}(q,\mathcal{E},\mathcal{M})
\]

We note that the task will not be used directly to query the bank. Instead, the agent generates queries from the given task for both question matching and experience-content matching. While existing systems retrieve through similarity matching~\cite{xu2025amemagenticmemoryllm,park2023generativeagentsinteractivesimulacra}) or active exploration strategies (e.g., in which the agent  probes the memory bank iteratively)~\cite{chhikara2025mem0buildingproductionreadyai,long2025seeinglisteningrememberingreasoning}, our \textbf{active retrieval} design enables the agent to retrieve relevant information from multiple dimensions, which contrasts with traditional search actions by allowing the agent to seek structural analogies or reasoning patterns rather than simple semantic overlaps, granting the agent higher autonomy in defining what constitutes "relevant" information for a complex forecasting query. We retrieve the top-k experiences. Each experience is ranked by the following score:
\[
Score = Weight * Sim(exp,query)
\]
When calculating the score, we not only consider the similarity between experiences and queries, but also multiply it by an experience weight that is updated during the evolution cycle.


\subsection{Compile}
The agent transforms retrieved experiences into task-adaptive guidance:
\[
g = \textsc{CompileGuideline}(q, E_q, \hat m).
\]
\textsc{CompileGuideline} operationalizes the role of the Guideline Bank: it selects and applies a meta-guideline $\hat m$ to turn the retrieved experience set $E_q$ into an executable, task-specific guideline $g$ for the current task $q$. Concretely, given $E_q$, \textsc{Live-Evo} performs meta-cognitive compilation by (i) extracting cross-experience regularities, (ii) grounding them in the current task context, and (iii) instantiating a guideline $g$ conditioned on $\hat m$ to steer downstream decision making. 

In contrast, prior approaches typically either concatenate retrieved logs as additional context or rely on fixed abstraction operators (e.g., summaries or heuristic rules) that remain static and do not improve from online feedback~\cite{wang2024agent,xu2025amemagenticmemoryllm}. 



\subsection{Act}
Conditioned on the task and the derived guideline, the agent executes a policy:
\[
r_q, \tau_q = \textsc{Act}(q \mid g),
\]
where $\tau_q$ denotes the trajectory, and $r_q$ denotes the resulting outcome signal. The structure of $r$ depends on the evaluation regime. In traditional reasoning or search benchmarks, $r$ is often binary, reflecting task success or failure. In contrast, online benchmarks (e.g.  \citet{yang2025llmasaprophetunderstandingpredictiveintelligence}) yield continuous feedback. These dense signals provide a richer learning substrate for memory evolution than sparse correctness-based rewards.

For every task, we additionally conduct a \emph{contrastive evaluation} to measure the causal impact of retrieved experience at action time. Concretely, we execute the agent again without the compiled guideline. We then compare the resulting outcomes to quantify whether memory usage provides a net benefit on that task. This comparison will later be used to update memory.


\subsection{Update}
Finally, the agent incorporates new experience into the memory bank. The update mechanism governs how experience accumulates over time and is grounded in objective environmental feedback. Concretely, from \emph{Contrastive Evaluation} we obtain the empirical gain of using the compiled guideline relative to the memory-free baseline. This gain is used to adjust the retrieval weights of the selected experiences: when the guideline improves performance, the corresponding experience weights are increased; when it harms performance, the weights are decreased. 
This reinforcement-and-decay dynamic is analogous to human memory, where useful experiences are strengthened through repeated success while misleading or outdated ones are gradually suppressed. In addition, failures trigger a reflection step that produces a new meta-guideline, which is added to the meta-guideline bank to improve future guideline compilation.

After processing a batch, we further perform selective experience acquisition rather than indiscriminately storing every trajectory. We identify the worst-performing fraction of tasks under the memory-on setting and generate a candidate experience by summarizing and reflecting on the stored trajectory. We then re-evaluate the task with this candidate experience, and commit it to the experience bank only if it yields a statistically significant improvement over the original memory-on score. This selective write-back controls memory growth while ensuring that new entries are justified by measurable gains.

\section{Experiment}

\subsection{Setup}  


We evaluate \textsc{Live-Evo} on \textbf{Prophet Arena}~\cite{yang2025llmasaprophetunderstandingpredictiveintelligence}, a future-prediction benchmark spanning the latest 10 weeks with 500 tasks in total. Each task contains a question, a candidate list, and a bid-price snapshot taken 6 hours before close, which we use to compute returns relative to market consensus. We enforce strict time-based retrieval on google-search tool to prevent information leakage past the close time. We also evaluate on \textbf{Xbench-DeepResearch}~\cite{chen2025xbenchtrackingagentsproductivity} to assess generalization beyond future prediction. We split the benchmark into 10 folds, learn experience sequentially across folds, and report the overall average accuracy.

We use GPT-4.1-mini as the backbone model for most experiments if not specified. All experiments use a \texttt{temperature}=0.2, with \texttt{bad\_case\_percentile}=0.3, \texttt{min\_brier\_improvement}=0.05, and \texttt{experience \_similarity\_threshold}=0.5.

\begin{table*}[ht!]
\small
\centering

\begin{tabular}{lccccccccccc}
\toprule
\textbf{Model} 
& \textbf{W1} & \textbf{W2} & \textbf{W3} & \textbf{W4} & \textbf{W5} 
& \textbf{W6} & \textbf{W7} & \textbf{W8} & \textbf{W9} & \textbf{W10} 
& \textbf{Avg} \\
\midrule

\multicolumn{12}{l}{\textbf{Base Model}} \\
GPT-4.1-mini & 0.18 & 0.20 & 0.31 & 0.18 & 0.24 & 0.26 & 0.23 & 0.25 & 0.23 & 0.15 & 0.22 \\
\midrule

\multicolumn{12}{l}{\textbf{Deep Research Methods}} \\

MiroFlow & 0.28 & 0.08 & 0.53 & 0.44 & 0.40 & 0.43 & 0.27 & 0.33 & 0.26 & 0.22 & 0.32 \\
Qwen Deep Research  & 0.17 & 0.22 & 0.23 & 0.15 & 0.22 & 0.22 & 0.21 & 0.19 & 0.20 & 0.13 & 0.20 \\

Live-Evo (w/o Experience) & 0.18 & 0.24 & 0.20 & 0.13 & 0.27 & 0.19 & 0.22 & 0.22 & 0.13 & 0.11 & 0.19 \\
\midrule


\multicolumn{12}{l}{\textbf{Self-Evolving Memory Methods}} \\
ReMem & 0.19 & 0.23 & \textbf{0.14 }& 0.11 & 0.21 & 0.18 & 0.19 & 0.17 & 0.15 & 0.11 & 0.16 \\

Live-Evo \textbf{(ours)} & 0.19 & 0.17 & 0.16 & \textbf{0.10} & \textbf{0.17} & \textbf{0.12} & 0.19 & \textbf{0.15} & \textbf{0.13} & \textbf{0.10} & \textbf{0.14} \\
\bottomrule
\end{tabular}
\caption{Brier Score (Lower the better) on Prophet-Arena - Weekly Performance Comparison }\
\label{tab:weekly_results_reversed}
\end{table*}

\begin{figure*}[h!] 
    \centering
    \begin{subfigure}[b]{0.48\linewidth}
        \centering
        \includegraphics[width=\linewidth]{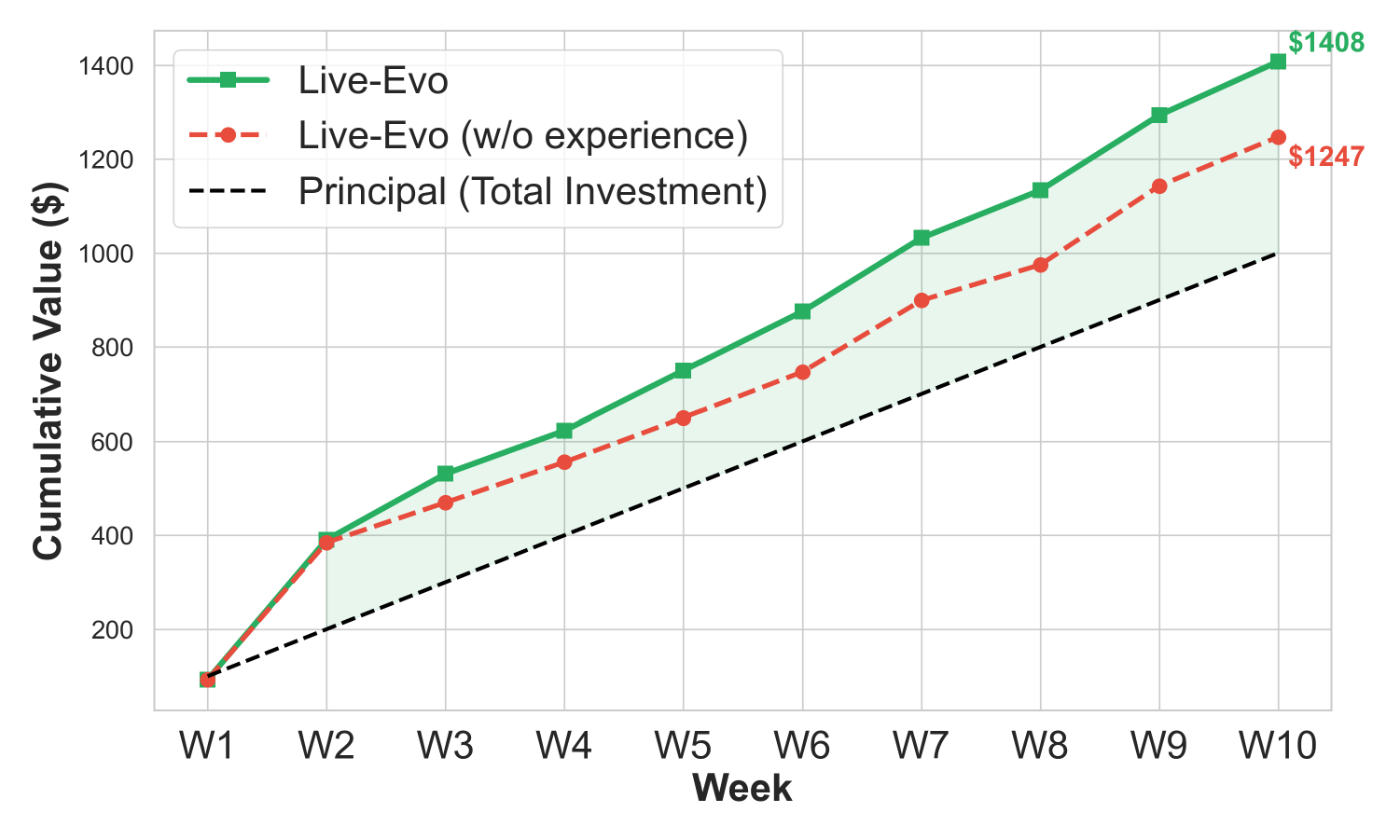}
        \caption{Cumulative Portfolio Value (Invest \$100 Per Week)}
        \label{fig:portfolio_value} 
    \end{subfigure}
    \hfill 
    \begin{subfigure}[b]{0.48\linewidth}
        \centering
        \includegraphics[width=\linewidth]{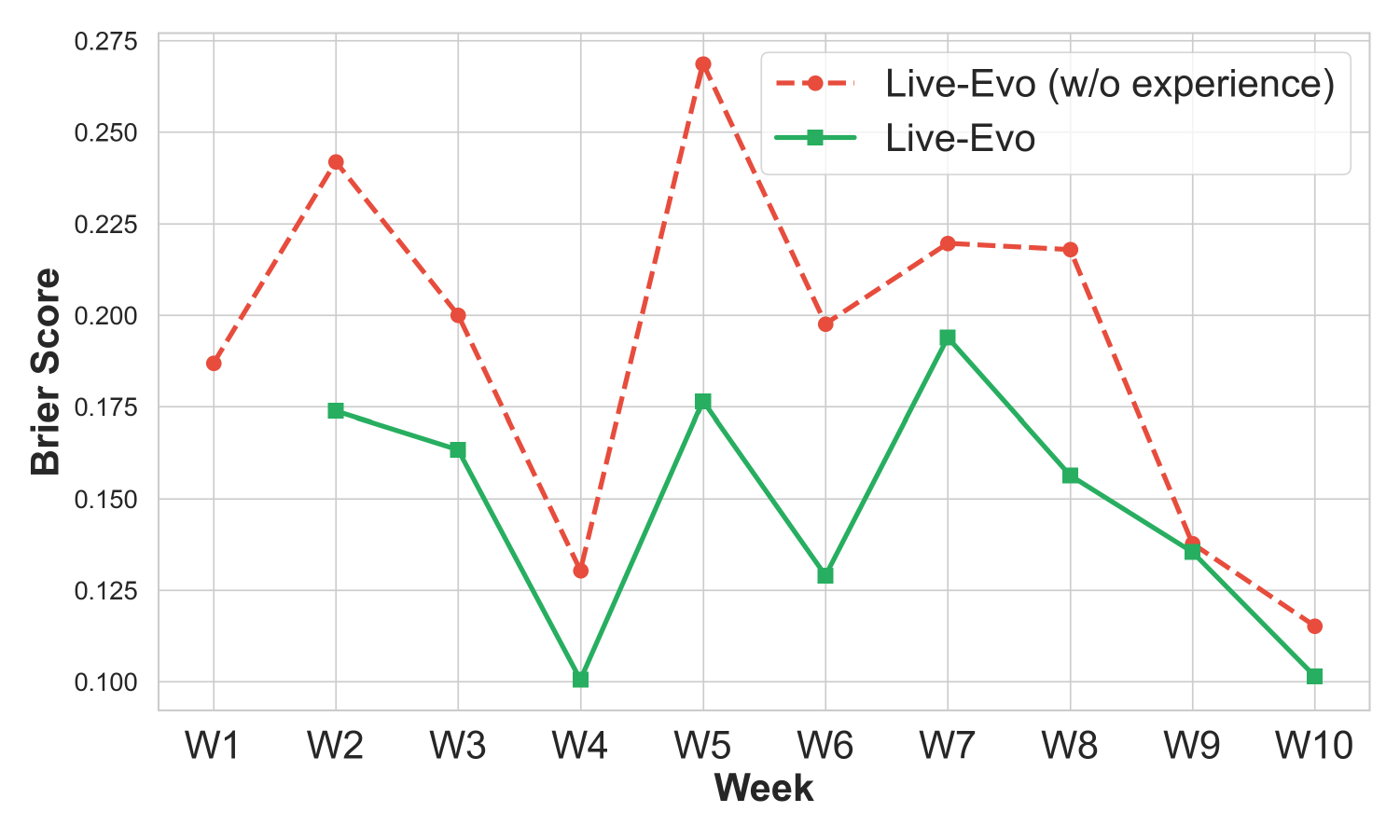}
        \caption{Brier Score Comparison}
        \label{fig:brier_score} 
    \end{subfigure}
    
    \caption{Performance Analysis Comparison. (a) shows the cumulative portfolio value, and (b) shows the Brier score comparison. }
    \label{fig:combined_analysis} 
\end{figure*}

\paragraph{Metrics.}  For XBench-DeepResearch, we use \textit{accuracy} as metrics.
For Prophet Arena, we use \textit{Brier Score} and \textit{Market Return} as metrics.
Given a query $q$ and a set of candidate outcomes
$\mathcal{C} = \{c_1, \ldots, c_K\}$, the agent outputs probabilities of each candidates
$\hat{\mathbf{p}} = (\hat{p}_1, \ldots, \hat{p}_K)$ over the outcomes.
Let $\mathbf{y} \in \{0,1\}^K$ denote the realized outcome and
$\mathbf{m} = (m_1, \ldots, m_K)$ the corresponding prediction market prices.
We report the multiclass Brier score as a calibration metric. A lower Brier score indicates that the predicted probabilities are closer to the true real-world outcomes.
\begin{equation}
\mathrm{BS} = \sum_{k=1}^K (\hat{p}_k - y_k)^2,
\end{equation}
We also compute the market return to quantify the advantage of \textsc{Live-Evo} over market-based baselines. The return is obtained by taking a unit long position on outcome $c_k$ whenever the predicted probability $\hat{p}_k$ exceeds the market-implied probability $m_k$:
\begin{equation}
R = \sum_{k=1}^K \mathbb{I}[\hat{p}_k > m_k]\,(y_k - m_k).
\end{equation}



\paragraph{Baselines.}
We compare \textsc{Live-Evo} with the following methods as baselines. \textbf{(1) Base Models} . We retrieve the top-10 web search results for each query and provide the model with a summarized version of these websites generated by the model itself. The model is then required to output the probability distribution based on this static information. \textbf{(2) Deep Research Methods.}  We evaluate two representative open-source frameworks, \texttt{Qwen Deep Research}~\cite{tongyidr} and \texttt{MiroFlow}~\cite{2025mirothinker} which support multiple tools and complex multi-agent workflows. We also evaluate \textsc{Live-Evo} without experience, representing the base search agent without evolution. 
\textbf{(3) Self-Evolving Memory Systems.} \textit{ReMem}~\cite{wei2025evomemorybenchmarkingllmagent}: a self-evolving agent baseline that constructs summarized experiences from raw trajectories and retrieves relevant memories at test time.


\subsection{Main Results}

Table~\ref{tab:weekly_results_reversed} reports the Brier scores over the most recent 10 weeks of the Prophet Arena benchmark. All methods use GPT-4.1-mini as the foundation model.
We compare our method against a Base Model and several open-source Deep Research frameworks.
We do not include closed-source Deep Research systems as baselines, because their search tools do not support strict time-based filtering. 

\begin{table*}[!ht]
\centering
\small
\caption{Generalization of \textsc{Live-Evo} across different foundation models. We report Brier score (lower is better) and cumulative market return (higher is better), along with relative improvements over the corresponding base agents.}
\label{tab:generalization_study}
\begin{tabular}{llcc|cc}
\toprule
\multirow{2}{*}{\textbf{Base Model}} & \multirow{2}{*}{\textbf{Method}} & \multicolumn{2}{c|}{\textbf{Brier Score ($\downarrow$)}} & \multicolumn{2}{c}{\textbf{Market Return ($\uparrow$)}} \\
\cmidrule(lr){3-4} \cmidrule(lr){5-6}
& & \textbf{Value} & \textbf{Imp.} & \textbf{Value} & \textbf{Imp.} \\
\midrule

\multirow{2}{*}{GPT-4.1-mini} 
& Base (w/o Mem) & 0.19 & -- & 1.24 & -- \\
& \textbf{Live-Evo } & \textbf{0.14} & \textit{20.8\%} & \textbf{1.46} & \textit{12.9\%} \\
\midrule

\multirow{2}{*}{GPT-4.1} 
& Base (w/o Mem) & 0.18 & -- & 1.13 & -- \\
& \textbf{Live-Evo} & \textbf{0.17} & \textit{3.0\%} &1.18 & \textit{4.4\%} \\
\midrule

\multirow{2}{*}{GPT-5-mini} 
& Base (w/o Mem) & 0.16 & -- & 1.34 & -- \\
& \textbf{Live-Evo} & \textbf{0.15} & \textit{4.5\%} & \textbf{1.36} & \textit{1.6\%} \\
\midrule

\multirow{2}{*}{Qwen3-8B} 
& Base (w/o Mem) & 0.19 & -- & 1.20 & -- \\
& \textbf{Live-Evo } & \textbf{0.18} & \textit{3.5\%} & \textbf{1.21} & \textit{0.5\%} \\
\bottomrule
\end{tabular}
\end{table*}

\paragraph{Result Analysis}
The results demonstrate that our agent achieves state-of-the-art performance in terms of the average Brier score, and outperforms all baselines in the majority of individual weeks.

Open-source Deep Research methods perform relatively poorly on this benchmark. This is expected, as they are optimized for discovering partial clues or supporting evidence, rather than producing calibrated probabilistic forecasts of future events. In practice, these methods are often misled by incomplete or temporally fragile signals.

The ReMem baseline shows a consistent improvement over the static Base Model (GPT-4.1-mini), indicating that incorporating self-evolving memory is beneficial for future prediction. However, its performance remains weaker than Live-Evo, highlighting the importance of actively managing and adapting experiences. These results confirm that our design more effectively leverages past experience under continuously evolving, real-world conditions.

\paragraph{Performance Comparison}

We compare \textsc{Live-Evo} with its underlying base search agent which isolates the contribution of the proposed experience management system.

Figure~\ref{fig:portfolio_value} illustrates the cumulative market returns under a simplified investment strategy. Assuming an investment of \$100 per week, \textsc{Live-Evo} achieves a \$150 higher return over the 10-week period. Notably, the performance gap between the two agents widens over time, indicating that \textsc{Live-Evo} continuously improves its decision quality as more experience is accumulated. Figure~\ref{fig:brier_score} further reports the weekly Brier scores. \textsc{Live-Evo} consistently outperforms the base agent across all weeks. The improvement is particularly pronounced during periods where the base agent exhibits poor calibration, such as Weeks 5 and 6. These results suggest that \textsc{Live-Evo} can stabilize predictions under difficult or volatile conditions.


\begin{table}[ht!]
\centering
\caption{Acc. on Xbench-DeepResearch. All methods are tested with GPT-4.1-mini.}
\label{tab:xbench}
\begin{tabular}{lccccccccccc}
\toprule
\textbf{Methods} & \textbf{Acc} \\
\midrule
Qwen-DeepResearch & 0.43 \\
MiroFlow & 0.45 \\
ReMem & 0.40 \\
\textbf{Live-Evo (ours)} & \textbf{0.46} \\
\bottomrule
\end{tabular}
\end{table}

\begin{table*}[ht!]
\centering
\small
\caption{Ablation Study Relative to the Full-Memory Model.
Color intensity indicates the magnitude of relative change compared to the full-memory setting.}
\label{tab:ablation_full_colored}
\begin{tabular}{lcccc}
\hline
\textbf{Setting} 
& \textbf{Avg. Brier} 
& \textbf{Change (\%)} $\downarrow$
& \textbf{Avg. Return} 
& \textbf{Change (\%)} $\uparrow$ \\
\hline
\textbf{Live-Evo}
& \textbf{ 0.14} & -- 
& \textbf{1.46} & -- \\


w/o weight-update
& 0.17 & \degrade{40} +17.01\%
& 1.34 & \degrade{20} -8.01\% \\

w/o meta-guideline
& 0.16 & \degrade{30} +10.88\%
& 1.41 & \degrade{10} -3.42\% \\

w/o guideline-compile
& 0.16 & \degrade{30} +11.56\%
& 1.16 & \degrade{50} -20.40\% \\

w/o active-retrieve
& 0.17 & \degrade{40} +14.97\%
& 1.22 & \degrade{40} -16.77\% \\
\hline
\end{tabular}
\end{table*}
\subsection{Additional Results with Different Models}

To evaluate the robustness of \textsc{Live-Evo} across foundation models of varying capacity and provenance, we conduct experiments on Prophet Arena with GPT-4.1-mini, GPT-4.1, GPT-5-mini, and Qwen3-8B, covering both closed-source and open-source models (See Table~\ref{tab:generalization_study}). For each model, we compare \textsc{Live-Evo} against its corresponding base agent without experience.

Across all evaluated models, \textsc{Live-Evo} consistently improves both Brier score and market return. These results demonstrate that the proposed experience management mechanism is broadly compatible with heterogeneous backbone models and does not rely on model-specific heuristics.

Notably, the largest relative improvement is observed with GPT-4.1-mini. This behavior is expected for two reasons. First, weaker base models exhibit greater headroom for improvement. Second, they generate more frequent failure cases during early weeks, which in turn provide richer supervisory signals for experience correction and guideline refinement. In contrast, stronger models such as GPT-5-mini already produce well-calibrated predictions, leaving less room for further gains.


\subsection{Results on Deep Research Benchmark}

Although \textsc{Live-Evo} is not specifically designed for traditional deep research tasks, it nevertheless demonstrates competitive and consistent advantages over both deep research frameworks and prior self-evolving memory methods.


\begin{figure*}[ht!]
    \centering
    \includegraphics[trim=1cm 7.5cm 1cm 1cm, clip, width=1\linewidth]{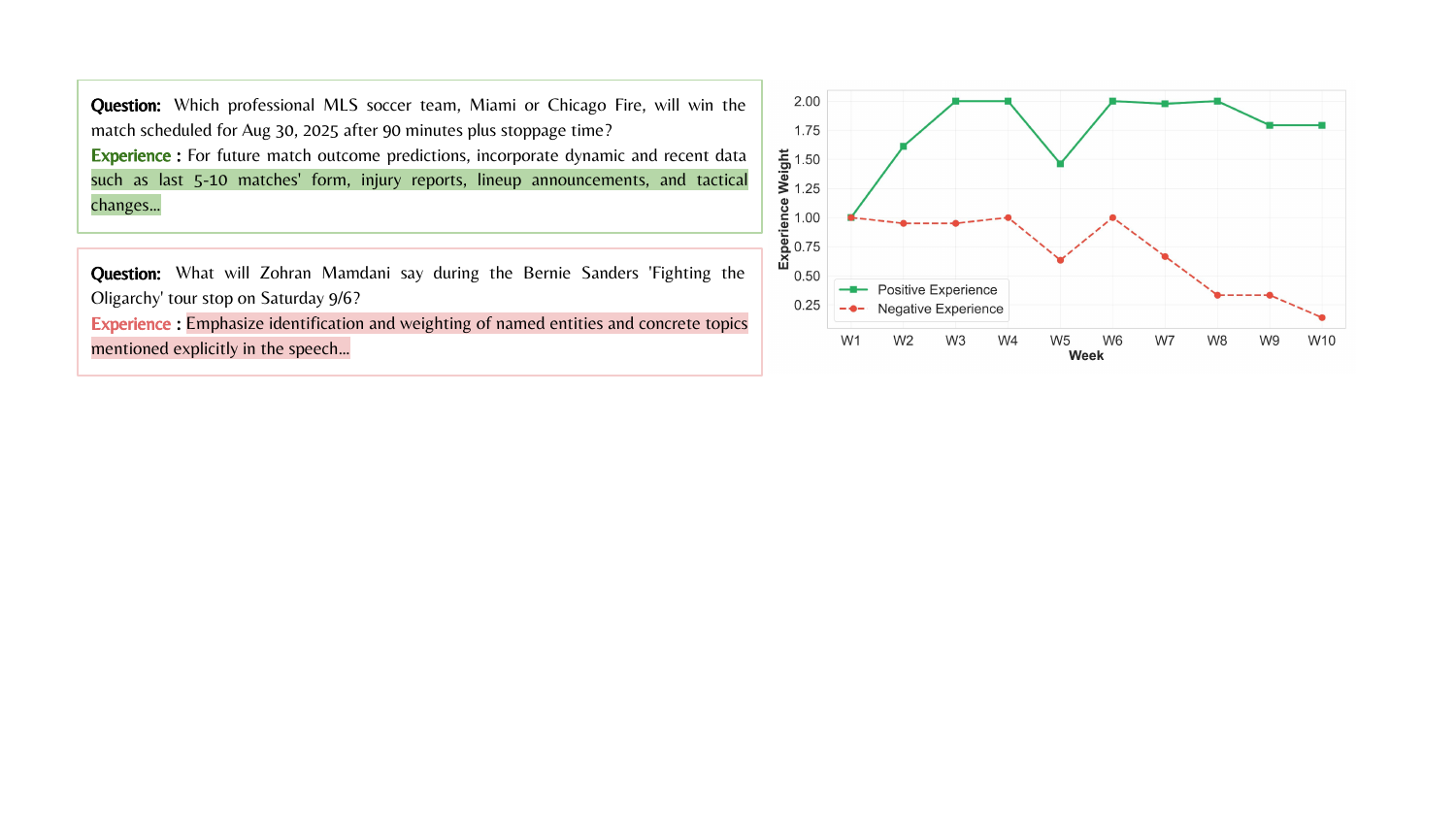}
    \caption{Case Study. The figure contrasts a high-weight experience (green), which provides reusable methods, with a low-weight experience (red), which contains hallucinations, and shows how their weights evolve weekly.}
    \label{fig:case_study}
\end{figure*}

As shown in Table~\ref{tab:xbench}, \textsc{Live-Evo} achieves the highest accuracy among all evaluated methods. Compared to specialized deep research systems such as Qwen-DeepResearch and MiroFlow, LiveEvo attains superior performance despite lacking task-specific heuristics for evidence exploration. This suggests that experience management learned under live and non-stationary conditions generalizes beyond future prediction, benefiting broader reasoning tasks.


\subsection{Ablation Study}


We conduct ablation studies to assess the contributions of key components in \textsc{Live-Evo} (Table~\ref{tab:ablation_full_colored}). Removing any single module consistently degrades performance in both Brier Score and market return, indicating that each component is non-trivial: w/o weight-update fixes all experience weights, w/o meta-guideline removes meta-guidance bank for guideline generation, w/o guideline-synthesis directly uses retrieved experiences, and w/o active-retrieve queries memory using only the question. Among these variants, disabling guideline synthesis causes the largest drop in market return, underscoring the importance of converting accumulated experience into actionable guidance, while removing active retrieval or adaptive weight updates also leads to substantial degradation. Overall, the results show that Live-Evo’s improvements stem from the \textbf{synergistic interaction} of its components rather than any single design choice.

\subsection{Case Study}
To illustrate how the weight-update mechanism operates in practice, we analyze experiences with the \textbf{lowest} and \textbf{highest} learned weights. In Figure~\ref{fig:case_study}, the red box highlights a low-weight experience that contains a clear hallucination: the experience suggests retrieving the content of a speech, whereas the task requires predicting the outcome of the speech. Such mismatches lead to consistently poor downstream performance and are therefore downweighted over time.

In contrast, the green box corresponds to a high-weight experience that provides a reusable and task-aligned guideline, recommending the analysis of recent match forms. This experience consistently supports accurate predictions and is thus reinforced by the weight-update mechanism.

This case study demonstrates that \textsc{Live-Evo} can progressively filter out low-quality or misleading experiences by reducing their weights, while amplifying high-quality, transferable experience. As a result, the agent learns to rely on increasingly reliable guidance, leading to improved future prediction performance.


\section{Conclusion}
We introduced Live-Evo, the first online-evolving agentic memory system specifically designed for benchmarks with continuous, real-world feedback. By employing a four-stage evolutionary loop:Retrieve, Compile, Act, and Update, the system dynamically learns to optimize how past experiences are transformed into task-adaptive guidance. Our evaluation on the Prophet Arena benchmark demonstrates that \textsc{Live-Evo} achieves significant improvement over strong baselines. Furthermore, the system exhibits robust generalization on deep-research benchmarks. These results underscore the vital role of feedback-driven experience management in building persistent, adaptive agentic systems for non-stationary environments.

\section{Limitations}
While \textsc{Live-Evo} achieves strong performance, its design introduces several potential constraints. First, its reliance on the dense environment feedback ensures robust calibration but may limit applicability in settings with sparse or subjective feedback. Second, the \textbf{Verify Before Update} protocol strictly admits new experiences only with statistically significant gains, which can delay the adoption of subtle or emerging heuristics.

\bibliography{custom}

\appendix

\section{Appendix}


\subsection{Implementation Details} 
\paragraph{Base Search Agent. }To isolate the efficacy of the Live-Evo, We build a simple search agent and equipped the agent with basic google-search and web fetch tool. We use serper api as the google-search api and apply the time filter by editing the queries. The max turns of one task is set to 20. For web content that exceed the agent's max sequence length, we call the llm to summarize the content. 

\paragraph{Retrieve. } We calculate the semantic similarity based on all-MiniLM-L6-v2 model from the sentence-transformers library. The system enforces a minimum weighted similarity threshold of $\tau = 0.3$. Only experiences has higher relativity will be retrieved. 

\paragraph{Experience Weight Update. }
Specifically, we update the experience weights according to the following formula:
\[ Weight_{new} = Weight_{old} + (score_{noexp} - score_{exp}) \]

\subsection{Example Case}
\label{sec:appendix_case_study}

In this section, we present a comprehensive execution trajectory of the \textsc{Live-Evo} system on a specific future prediction task from the \textit{Prophet Arena} benchmark. This case study illustrates how the agent retrieves historical failures, synthesizes a dynamic guideline, executes actions based on that guideline, and achieves a superior Brier score compared to the baseline.

\subsubsection{Task Definition}
The agent is presented with a binary prediction task regarding an NFL game.

\begin{tcolorbox}[title=\textbf{Task Input}, colback=gray!5, colframe=black!75]
\textbf{Question:} Which professional football team, Cincinnati or Pittsburgh, will win the game scheduled for Oct 16, 2025? \\
\textbf{Ground Truth:} Cincinnati (1), Pittsburgh (0)
\end{tcolorbox}

\subsubsection{Phase 1: Retrieve}
Upon receiving the task, the agent queries the Experience Bank ($\mathcal{E}$). The system retrieves relevant past experiences where the agent previously failed due to over-reliance on betting odds or missed schedule changes. Two representative retrieved experiences are shown below:

\begin{table*}[ht!]
\centering
\caption{Performance Comparison: Baseline vs. Live-Evo}
\label{tab:comparison}
\begin{tabular}{p{0.45\linewidth} | p{0.45\linewidth}}
\hline
\textbf{Baseline Agent (No Memory)} & \textbf{Live-Evo Agent (With Memory)} \\ \hline
\textbf{Prediction:} Pittsburgh (0.65), Cincinnati (0.35) & \textbf{Prediction:} Cincinnati (0.55), Pittsburgh (0.45) \\
\textbf{Reasoning:} Relied on Pittsburgh's 4-1 record and winning streak. Heavily weighted betting odds favoring Steelers. & \textbf{Reasoning:} Identified close 33-31 victory conditions. Weighed resilience despite injuries and home advantage more heavily than static odds. \\
\textbf{Brier Score:} 0.5329 & \textbf{Brier Score:} 0.2500 \\
\hline
\end{tabular}
\end{table*}

\begin{tcolorbox}[title=\textbf{Example Retrieved Experiences},breakable, colback=blue!5, colframe=blue!75]
\small
\textbf{Experience (Sports/NCAAF)} \\
\textit{Failure Reason:} The agent over-relied on pre-game betting odds and recent season trends without accounting for roster changes or "home advantage" dynamics closer to the game date. \\
\textit{Improvement:} Incorporate dynamic, up-to-date info (roster, coaching) as the event approaches. Avoid static betting odds.
\textbf{Experience (Sports/MLS)} \\
\textit{Failure Reason:} The agent failed to update its prediction to reflect the \textbf{rescheduling} of the match, basing probabilities on outdated timing. \\
\textit{Improvement:} Always verify the event date and confirm the prediction is relative to the current schedule.
\end{tcolorbox}
One example meta guideline is: 
\begin{tcolorbox}[title=\textbf{Example Retrieved Experiences},breakable, colback=blue!5, colframe=blue!75]
\small
\textbf{Experience (Sports/NCAAF)} \\
\textit{Failure Reason:} Over-generalization of domain-specific lessons across fundamentally different task types and contexts. The synthesis process failed by inadequately discriminating between domain and task relevance, leading to the inappropriate transfer of detailed lessons from sports prediction experiences to a political individual behavior prediction task, and by insufficiently validating contextual alignment and specificity of lessons before integration. \\
\textit{Synthesis Instruction:} When generating guidelines from past experiences, explicitly verify that the domain, task type, and contextual factors closely align before transferring lessons; avoid importing detailed procedural or content-specific insights from experiences that differ substantially in domain or prediction target, instead extracting only high-level, abstract methodological principles with caution. Before incorporating lessons from past experiences, systematically assess domain congruence, task similarity, and context relevance; if experiences differ in domain or task type, restrict transferred lessons to broadly applicable methodological insights only, and clearly flag any assumptions or limitations in applicability within the guideline.
\end{tcolorbox}
\subsubsection{Phase 2: Compile}
Using the retrieved experiences and the meta-guideline, the agent synthesizes a task-specific guideline ($\mathcal{G}$) via the \textsc{Reuse} operator. This guideline explicitly warns against the specific pitfalls identified in the retrieved memory (e.g., verifying dates and ignoring early odds).

\begin{tcolorbox}[title=\textbf{Synthesized Guideline},breakable, colback=green!5, colframe=green!60!black]
\begin{itemize}
    \item \textbf{Dynamic Information:} Prioritize authoritative sources (e.g., official injury reports, press releases) close to the game date over early betting odds or historical reputation.
    \item \textbf{Schedule Verification:} Implement a workflow step to confirm the exact game date and update contextual data to avoid outdated inputs (derived from Exp 50fe0d0c).
    \item \textbf{Scenario Analysis:} Explicitly model the impact of key player absences (e.g., injury reports) and home vs. away advantages.
\end{itemize}
\end{tcolorbox}

\subsubsection{Phase 3: Act}
Guided by the synthesized instructions, the agent executes a search strategy. Unlike a standard agent that might immediately look up win probabilities, the \textsc{Live-Evo} agent follows the guideline to first verify the schedule and check specific injury reports.

\begin{tcolorbox}[title=\textbf{Agent Execution Trajectory (Abbreviated)},breakable, colback=white, colframe=gray]
\small
\textbf{Action 1: Web Search} \\
\texttt{Query: "Cincinnati vs Pittsburgh NFL game Oct 16 2025 injury report roster updates"}

\textbf{Observation 1 (Search Results):} 
Results indicate the game date matches the query. Snippets show specific injury reports: "Bengals TE Mike Gesicki removed from injury report," "Bengals DE Trey Hendrickson ruled OUT."

\textbf{Action 2: Visit Page} \\
\textbf{Observation 2 (Web Content):}
\textit{[Content summarized for brevity]}
Detailed tables confirm:
\begin{itemize}[nosep]
    \item \textbf{Cincinnati:} Trey Hendrickson (DE) is OUT; Tanner Hudson (TE) is OUT.
    \item \textbf{Pittsburgh:} Calvin Austin III (WR) is OUT; Miles Killebrew (S) is OUT.
\end{itemize}
Date confirmation: Thursday, Oct 16, 2025.

\textbf{Action 3: Web Search} \\
\texttt{Query: "Cincinnati Bengals vs Pittsburgh Steelers recent performance 2025 season stats"}

\textbf{Observation 3 (Search Results):} 
Results confirm the game conclusion
\end{tcolorbox}

\subsubsection{Phase 4: Result \& Update (Evaluation)}

The agent synthesizes the gathered evidence. \textbf{A detailed comparison between the baseline and the \textsc{Live-Evo} agent is presented in Table~\ref{tab:comparison}.} While the Baseline agent (without memory) relied on Pittsburgh's superior record (4-1) and betting odds, the \textsc{Live-Evo} agent incorporated the specific game-day dynamics and injury resilience found during the guided search.

The \textsc{Live-Evo} system achieved a \textbf{Brier Score improvement of 0.2829}. Following this success, the weight of the retrieved experiences is increased, reinforcing the guideline to "verify schedule" and "ignore early odds" for future sports prediction tasks.

\subsection{Prompts}
Prompt: Retrieve Query Generation shows the prompt that guide the agent to generate retrieve queries for the experience bank and meta-guideline bank. 
Prompt: Guideline Compile shows the prompt that guide the agent generate guideline based on experiences, meta-guideline and current tasks.
Prompt: Base Agent Prediction shows how the base search agent will act given the task and the guideline.

\begin{tcolorbox}[title=\textbf{Prompt: Search Query Generation}, colback=gray!5, colframe=black!75,label=box:searchquery]
You are exploring an experience database to find relevant past predictions that can help with a new task.

\vspace{0.5em}
\textbf{Current Task:} Which professional football team, Cincinnati or Pittsburgh, will win the game scheduled for Oct 16, 2025?

\vspace{0.5em}
The experience database contains past prediction experiences with these fields:
\begin{itemize}
    \item \texttt{question}: The prediction question/task title
    \item \texttt{improvement}: Key insights on how to improve similar predictions
    \item \texttt{failure\_reason}: What went wrong in past predictions
    \item \texttt{missed\_information}: Information sources that were missed
    \item \texttt{category}: Domain category (politics, technology, etc.)
\end{itemize}

Generate 2-3 search queries. For each, specify the text and search type: ``question'' or ``experience''.

\vspace{0.5em}
\textbf{Output as JSON:}
\begin{small}
\begin{verbatim}
{
  "queries": [
    {"query": "...", "search_target": "..."}
  ]
}
\end{verbatim}
\end{small}
\end{tcolorbox}

\begin{tcolorbox}[title=\textbf{Prompt: Guideline Compile}, colback=gray!5, colframe=black!75,label=box:guideline]
You are synthesizing insights from past prediction experiences to create a guideline for a new prediction task.

\textbf{Current Task:} Which professional football team, Cincinnati or Pittsburgh, will win the game scheduled for Oct 16, 2025?

\textbf{Relevant Past Experiences Found:}
[Experience 1 Summary: Over-relied on betting odds, missed key defensive injuries...]
[Experience 2 Summary: Over-relied on historical trends, missed roster changes...]

\textbf{CRITICAL: Experience Applicability Check}
Before applying lessons from past experiences, you MUST assess whether each experience is truly applicable to this specific task type. Identify which lessons are directly applicable vs. need adaptation.

Based on these experiences AND the applicability check, generate a FOCUSED and ACTIONABLE guideline (3-5 bullet points) for this prediction task. 

\textbf{Output ONLY the bullet points.}
\end{tcolorbox}

\begin{tcolorbox}[
    title=\textbf{Prompt: Base Agent Prediction}, 
    colback=gray!5, 
    colframe=black!75,
    fonttitle=\sffamily,
    label=box:baseagent
]
You are tasked with predicting the probability of different outcomes for the following event:

\medskip
\textbf{Event:} Which professional football team, Cincinnati or Pittsburgh, will win the game scheduled for Oct 16, 2025? \\
\textbf{Possible outcomes:} ``Cincinnati'', ``Pittsburgh''

\medskip
\textbf{\textsf{Task-Specific Guideline}}
\begin{itemize}[leftmargin=1.5em, nosep]
    \item Incorporate dynamic, up-to-date information close to the game date, including granular injury reports (e.g., Trey Hendrickson status).
    \item Avoid over-reliance on early betting odds or historical reputation.
    \item Explicitly model and quantify the impact of key player absences and home-field advantage.
\end{itemize}

\medskip
\textbf{\textsf{CRITICAL: How to Properly Use This Guideline}}
\begin{enumerate}[leftmargin=1.5em, nosep]
    \item \textbf{Verify Applicability}: Assess if the current task matches the context of the lesson.
    \item \textbf{Trust Your Current Research}: If fresh findings contradict the guideline, prioritize fresh evidence.
\end{enumerate}

\tcblower 

\textbf{Your task:}
\begin{enumerate}[nosep]
    \item Research this event by searching for relevant information online.
    \item Analyze the information to assess the likelihood of each outcome.
    \item Provide a probability estimate (between 0 and 1).
\end{enumerate}

\medskip
\textbf{Output as JSON.}
\end{tcolorbox}
\end{document}